\documentclass{article}
\usepackage{spconf,amsmath,graphicx,cite,url}

\title{IMPROVING PROSODY FOR UNSEEN TEXTS IN SPEECH SYNTHESIS BY UTILIZING LINGUISTIC INFORMATION AND NOISY DATA}

\name{Zhu Li, Yuqing Zhang, Mengxi Nie, Ming Yan, Mengnan He, Ruixiong Zhang, Caixia Gong}
\address{DiDi Chuxing, Beijing, China}

\begin{document}
\maketitle
\begin{abstract}
Recent advancements in end-to-end speech synthesis have made it possible to generate highly natural speech. However, training these models typically requires a large amount of high-fidelity speech data, and for unseen texts, the prosody of synthesized speech is relatively unnatural. To address these issues, we propose to combine a fine-tuned BERT-based front-end with a pre-trained FastSpeech2-based acoustic model to improve prosody modeling. The pre-trained BERT is fine-tuned on the polyphone disambiguation task, the joint Chinese word segmentation (CWS) and part-of-speech (POS) tagging task, and the prosody structure prediction (PSP) task in a multi-task learning framework. FastSpeech 2 is pre-trained on large-scale external data that are noisy but easier to obtain. Experimental results show that both the fine-tuned BERT model and the pre-trained FastSpeech 2 can improve prosody, especially for those structurally complex sentences.
\end{abstract}
\section{Introduction}
\label{sec:intro}

Generating highly natural speech has been made possible by recent advances in end-to-end text-to-speech (TTS) synthesis \cite{shen2018natural, shen2020non, ren2019fastspeech, ren2020fastspeech}. However, for unseen texts, the prosody (especially pitch and rhythm) of synthesized speech is relatively unnatural (e.g., less dynamic pitch variation, inappropriate phrasing, and improper breaks). For example, in out-of-set speech synthesis for language like Mandarin, which has no blank space between neighboring words, occasional word segmentation errors or poor phoneme duration prediction often leads to improper pauses between characters of a word, which might hinder listeners' comprehension.

To improve the modeling of prosody, many recent studies use linguistic or BERT-derived features to provide additional contextual information to TTS models \cite{hayashi2019pre, fang2019towards, zhu2019probing, zhang2020unified, kenter2020improving, xiao2020improving, xu2021improving, chien2021hierarchical, jia2021png}. For example, researchers have utilized character \cite{zhu2019probing, xiao2020improving}, sub-word \cite{hayashi2019pre, kenter2020improving}, word \cite{chien2021hierarchical}, or sentence embeddings \cite{hayashi2019pre, xu2021improving} derived from a pre-trained BERT model for augmenting phoneme embeddings in end-to-end TTS systems. However, these experiments are mostly based on attention-based TTS models like Tacotron 2 \cite{hayashi2019pre, zhu2019probing, xiao2020improving, xu2021improving}. Relatively few studies have investigated incorporating linguistic knowledge into duration-based acoustic models like FastSpeech 2 \cite{chien2021hierarchical}. In addition, the overall improvements are sometimes rather limited \cite{fang2019towards}, and there are few studies on enhancing BERT to improve the performance of TTS synthesis tasks further.


Generating natural prosody relies not only on rich linguistic features but also on high-performance and robust acoustic models. For duration-based models like FastSpeech 2, which leverages an explicit duration predictor, low phoneme duration prediction accuracy often results in unnatural pronunciation or wrong placement of breaks. In addition, training these acoustic models often requires a sizable set of high-quality speech data. Most prior studies have focused on how to get the reliable duration label to train the duration predictor and how to optimize the duration prediction in an end-to-end manner (e.g., \cite{donahue2020end}), but relatively few have developed the idea of utilizing external data to pre-train the duration predictor and enhance its generalization ability. 

The main contributions of this work are as follows:

1) We introduce an all-inclusive fine-tuned BERT-based front-end, which provides linguistic features that contain rich linguistic information by combining phoneme embeddings with character embeddings. We employ the multi-task learning technique to fine-tune the pre-trained BERT, with the aim of capturing task-specific linguistic knowledge. The subjective evaluation results show that incorporating this fine-tuned BERT-based front-end can improve prosody modeling.

2) We use the fine-tuned BERT for polyphone disambiguation in the grapheme-to-phoneme (G2P) conversion module, which simplifies the TTS front-end pipeline. 

3) We pre-train the encoder and the duration predictor of FastSpeech 2 using large-scale external data that are noisy but easier to obtain. Experimental results indicate that model pre-training based on the noisy AISHELL-1 dataset is effective in improving the prosody within the FastSpeech2-based TTS framework. Sound demos are available online.$\footnote{https://cookingbear.github.io/research/publications/finetuned-bert/}$


\section{Proposed Approach}
\label{sec:model}

\begin{figure*}[htb]
  \centering
    \includegraphics[width=140mm]{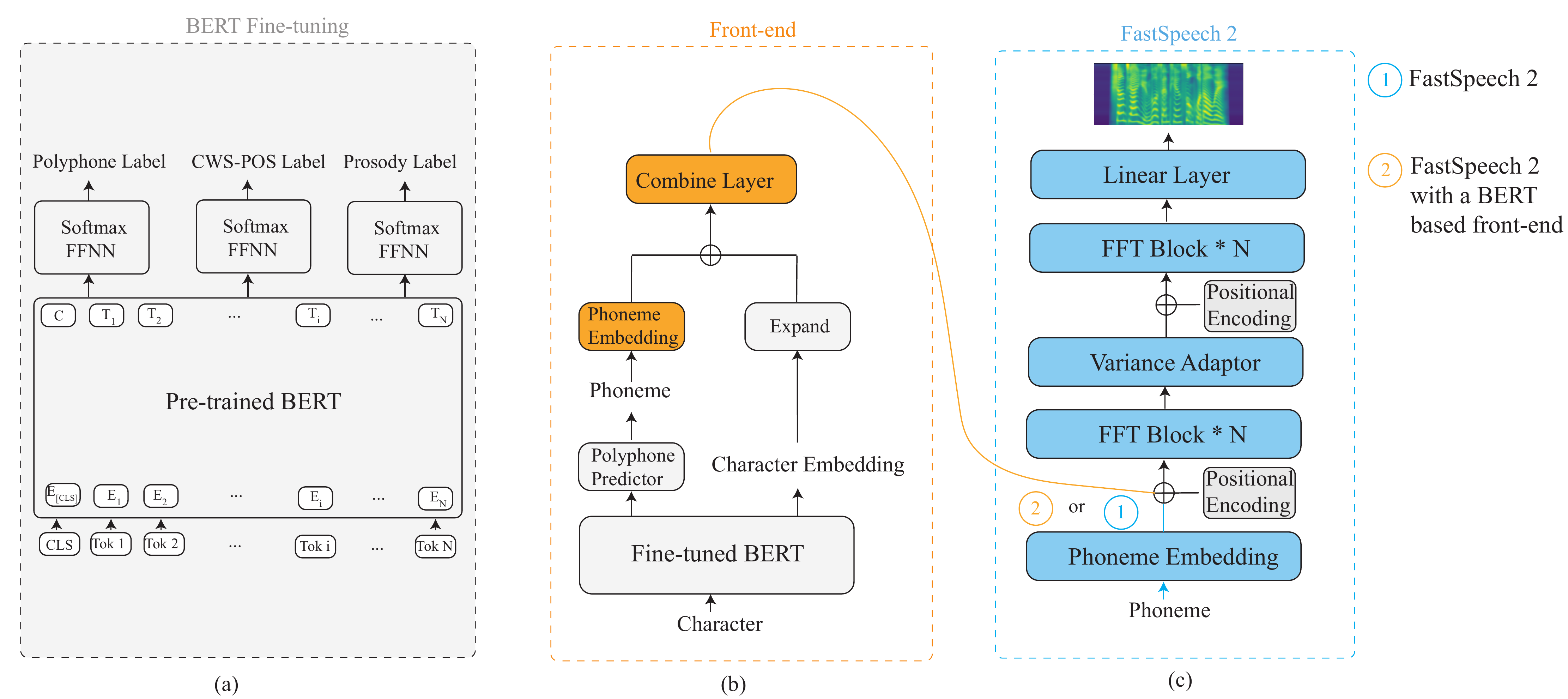}
\caption{The overall model architecture. (a) Overall procedures for fine-tuning BERT on different tasks. (b) The front-end based on a fine-tuned BERT model. (c) FastSpeech 2 with output features from the BERT-based front-end as input (orange line).}
\label{fig:model}
\end{figure*}

The proposed model architecture follows the original FastSpeech 2 \cite{ren2020fastspeech} and the BERT architecture \cite{devlin2018bert}. It consists of a TTS front-end based on a fine-tuned BERT model, and an acoustic model based on FastSpeech 2 with a pre-trained duration predictor, as illustrated in Figure \ref{fig:model}. 


\subsection{Fine-tuned BERT-based front-end}


\subsubsection{Fine-tuning BERT for linguistic information}

Pre-training provides task-independent general knowledge, and fine-tuning can presumably teach the model to rely more on representations useful for the task at hand. For instance, BERT embeddings fine-tuned on the joint CWS and POS tagging task might acquire more knowledge about phrasing. Therefore, if the fine-tuned embeddings are used in TTS synthesis, the phrasing pattern of generated speech for sentences that are hard to segment is likely to improve. In addition, multi-task learning has demonstrated its effectiveness in sharing the knowledge obtained from several related supervised tasks. Thus with multi-task fine-tuning, BERT representations can gain richer linguistic or contextual knowledge.

During the fine-tuning process, BERT takes the final hidden vector h for the $i^{th}$ input token as the representation of the character $Tok\ i$. A simple FFNN (Feed-Forward Neural Work) with a softmax classifier is added to the top of BERT to predict the probability of label $c$: 
\begin{equation}
   p(c|h) = softmax(W h) 
\end{equation}
where $W$ is the task-specific parameter matrix. We fine-tune all the parameters from BERT as well as $W$ jointly by maximizing the log-probability of the correct label. Specifically, we conduct single-task fine-tuning (BERT-Polyphone; BERT-CWS+POS; BERT-Prosody) and multi-task fine-tuning (BERT-Multi) based on the original character-based Chinese BERT-base model \cite{devlin2018bert}, which contains an encoder with 12 Transformer blocks, 12 self-attention heads, and a hidden size of 768.

\textbf{BERT-Polyphone}: BERT encoder with a prediction layer of the polyphone disambiguation task.

\textbf{BERT-CWS+POS}: BERT encoder with a prediction layer of the joint CWS and POS tagging task.

\textbf{BERT-Prosody}: BERT encoder with a predictor layer of the prosody structure prediction task.

\textbf{BERT-Multi}: BERT encoder with three prediction layers attached in parallel. BERT embeddings are fine-tuned in a multi-task learning framework.

\subsubsection{An all-inclusive front-end}

We propose a novel fine-tuned BERT-based front-end, which not only performs G2P conversion but also provides linguistic features that contain rich contextual information by combining phoneme embeddings with character embeddings. Specifically, the fine-tuned BERT helps determine the pronunciation of polyphonic characters in the G2P conversion module. Then the converted phonemes are sent to an embedding layer to obtain the phoneme embeddings. Meanwhile, from the fine-tuned BERT we extract the character embeddings, which are then upsampled to match the length of the phonemic symbols of each character. Finally, the character embeddings and the phonemic embeddings are concatenated as the input features to the subsequent layers. 


\subsection{Pre-trained FastSpeech2-based acoustic model}
\label{acoustic_mod}
Duration-based acoustic models commonly use external or internal alignment tools for obtaining ground-truth values of phoneme duration as targets to train the duration predictor. However, high-quality speech data are expensive and time-consuming to collect. We propose to lower acoustic models' reliance on high-quality data and enhance their generalization ability on out-of-set speech synthesis by using noisy data to pre-train the duration predictor. To verify its effectiveness in improving the prosody of synthesized speech for unseen texts, we pre-train the duration predictor on a large-scale noisy dataset and on a relatively small clean dataset, to see if they can achieve comparable performance. Moreover, inspired by the BERT pre-training mechanism \cite{devlin2018bert}, while pre-training the duration predictor, we randomly mask phonemes to pre-train the encoder (i.e., the first stack of FFT blocks) in Fastspeech 2, which acts as a data augmentation effect for pre-training the duration predictor. 



\begin{figure}[htb]
  \centering
    \includegraphics[width=60mm]{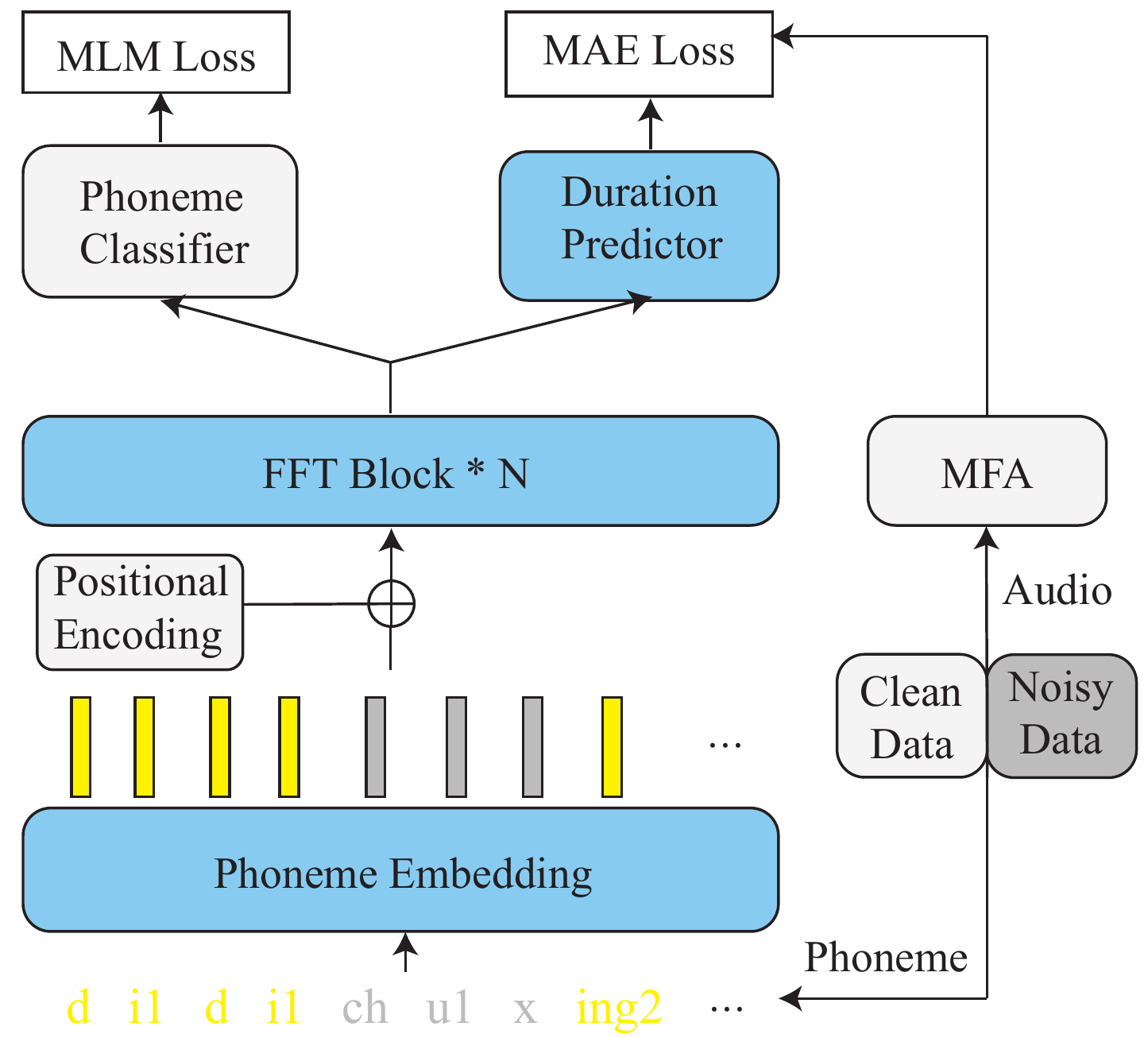}
\caption{Pre-trained duration predictor. The phonemes ``ch", ``u1" and ``x", and their embeddings are shaded in grey, indicating that they are masked.}
\label{fig:pretrained_dur}
\end{figure}

As shown in Figure \ref{fig:pretrained_dur}, the pre-training steps can be summarized as follows: We use the Montreal Forced Aligner (MFA) \cite{mcauliffe2017montreal} to extract the ground-truth duration of each phoneme. The phoneme sequences with some phonemes being randomly masked are sent into the phoneme embedding layer. Finally, the MAE (mean absolute error) loss and the MLM (masked language model) loss can be computed, and the total loss back-propagates through the network.



 

\section{Experiments}
\label{sec:exp}

\subsection{Datasets}

\subsubsection{Data for training acoustic models}
The training data used for this study is the Chinese Standard Mandarin Speech Corpus (CSMSC) \cite{csmsc2017}. CSMSC has 10,000 recorded sentences read by a female speaker, with the total audio length of about 12 hours of natural speech. We randomly split the dataset into two parts: 9500 samples for training and 500 samples for testing. The clean data used for pre-training the duration predictor is the AISHELL-3 training set (around 60 hours long) \cite{shi2020aishell}, which is a high-fidelity multi-speaker Mandarin speech corpus. And the AISHELL-1 training set (around 150 hours) \cite{bu2017aishell}, which is primarily used for building speech recognition systems, is selected as the noisy data to pre-train the duration predictor. 

\subsubsection{Data for fine-tuning BERT}

We fine-tuned BERT on a text corpus mined from the Chinese Wikipedia, containing 81108 sentences. The corpus was manually annotated by trained linguists. For the joint segmentation and tagging task, 35 tags for word classes and 2 tags for a character's position within a word were used, totaling 70 categories. For the prosody prediction task, three classes were used for the prosodic break of prosodic word (PW), prosodic phrase (PPH) and intonational phrase (IPH), respectively. The corpus was randomly partitioned into training and test sets with 71853 and 9255 sentences, respectively.

\subsection{Model Configuration} 
The configuration and hyper-parameters of the baseline model follow the original FastSpeech 2 implementation \cite{ren2020fastspeech}. It consists of 4 feed-forward Transformer (FFT) blocks in the encoder and the mel-spectrogram decoder. In each FFT block, the dimension of phoneme embeddings and the hidden size of the self-attention are set to 256. The combine layer is a 1D-convolutional network with a ReLU activation function, an input size of 1024, and output size 256. The output linear layer in the decoder converts the hidden states into 80-dimensional mel-spectrograms. The model is optimized with MAE. Generated mel-spectrograms are converted into waveforms using the multi-band MelGAN (MB-MelGAN) vocoder \cite{yang2021multi}. 


\textbf{Training details:} The duration predictors are pre-trained with the Adam optimizer (with $\epsilon$ = $10^{-9}$, $\beta_{1}$ = 0.9, $\beta_{2}$ = 0.98) for 10k steps with a batch size of 48, and the learning rate schedule proposed in \cite{vaswani2017attention} is applied. All acoustic models are trained with the Adam optimizer (with $\epsilon$ = $10^{-9}$, $\beta_{1}$ = 0.9, $\beta_{2}$ = 0.98) for 20K steps with a batch size of 48.

\subsection{Fine-tuned BERT performance}
\label{ssec:subhead1}
The performance of single-task fine-tuning and multi-task fine-tuning based on BERT is summarized in Table \ref{tab:my_label}. BERT-Multi slightly outperforms other single-task fine-tuning systems in terms of polyphone disambiguation and prosody prediction, except for the segmentation and tagging task. All fine-tuned systems achieve fairly good results on all tasks.

\begin{table}[!htbp]
    \caption{Performance of different systems.}
    \label{tab:my_label}
    \centering
      \resizebox{\columnwidth}{!}{
    \begin{tabular}{cccccc}
        \hline
        \textbf{Systems} & \textbf{ACC-Polyphone} & \textbf{F1-CWS+POS} & \multicolumn{3}{c}{\textbf{F1-Prosody}} \\
        & & & \textbf{PW} & \textbf{PPH} & \textbf{IPH} \\  \hline
        BERT-Polyphone & 98.47 & -  & - & - & - \\
        BERT-CWS+POS & - & 93.69 & - & - & - \\
        BERT-Prosody & - & - & 84.78 & 70.44 & 90.84 \\
        BERT-Multi & 98.86 & 93.34 & 85.06 & 71.90 & 91.34 \\
    \hline
    \end{tabular}}
\end{table}

\subsection{TTS performance}
\label{ssec:subhead2}

\subsubsection{Ablation studies}
\label{sssec:subsubhead01}

We conduct ablation studies to evaluate the effectiveness of combining a fine-tuned BERT-based front-end with a pre-trained FastSpeech2-based acoustic model in improving the prosody of synthesized speech. The overall speech quality is evaluated on the CSMSC test set using the mean opinion score (MOS) \cite{chu2006objective}. Ten native speakers were asked to rate the overall prosodic appropriateness of the synthesized speech. We compare the MOS of audio samples generated by our proposed model with other systems, including 1) GT, the ground-truth audios; 2) GT (Mel + MB-MelGAN); 3) FastSpeech 2; 4) w/ BERT: a BERT-based front-end;  5) w/ BERT-Multi: a fine-tuned BERT-based front-end. 6) and 7): clean/noisy means the pre-trained duration predictor with AISHELL3/AISHELL1 dataset. 8) w/ BERT-Multi + noisy.


\begin{table}[htb]
    \caption{The MOS with 95\% confidence intervals.}
    \label{tab:my_label2}
    \centering
    \resizebox{\columnwidth}{!}{
    \begin{tabular}{ll}
        \hline
        \textbf{System} & \textbf{MOS} \\
        \hline
        GT & $4.25 \pm 0.07$ \\ 
        GT (Mel + MB-MelGAN) &  $3.98 \pm 0.08$ \\ \hline
        FastSpeech 2 (Mel + MB-MelGAN) &  $3.74 \pm 0.07$\\
        \quad w/ BERT (Mel + MB-MelGAN) & $3.82 \pm 0.07$ \\
        \quad w/ BERT-Multi (Mel + MB-MelGAN) & $3.88 \pm 0.08$  \\
        \quad w/ clean (Mel + MB-MelGAN) & $3.78 \pm 0.07$ \\
        \quad w/ noisy (Mel + MB-MelGAN) & $3.80 \pm 0.08$ \\
        \quad w/ BERT-Multi + noisy (Mel + MB-MelGAN) & $3.90 \pm 0.09$ \\
    \hline
    \end{tabular}}
\end{table}

The results suggest that using a fine-tuned BERT-based front-end and a pre-trained FastSpeech2-based acoustic model can improve the prosody (with 0.14 and 0.06 MOS gains, respectively), as shown in Table \ref{tab:my_label2}. In addition, the w/ clean model and the w/ noisy model achieve comparable performance, suggesting that pre-training the encoder and the duration predictor on noisy data is feasible and can effectively improve the data efficiency and prosody at the same time. 


\subsubsection{Pairwise subjective scores}
We further conduct an ABX test to compare the models pairwisely. Synthesized audio samples from two different systems were presented to the same ten raters. And they were asked to select the one with more appropriate prosody. The evaluation text set contains 150 sentences and each sentence has more than 10 characters. These texts are randomly selected from the test set of the fine-tuning BERT experiments.

The results are summarized in Table \ref{tab:my_label3}. All fine-tuned BERT-based models outperform the models without fine-tuning and listeners preferentially choose the audios synthesized by BERT-Multi, suggesting that rich contextual information contained in fine-tuned BERT representations can be injected into the model. The w/ clean and the w/ noisy model perform fairly close, which suggests that pre-training on noisy data or clean data can bring about comparable improvements.





\begin{table}[htb]
    \caption{The subjective listening preference rate.}
    \label{tab:my_label3}
    \centering
    \resizebox{\columnwidth}{!}{
\begin{tabular}{llll}
    \hline
\textbf{ABX Choices} & \textbf{Former} & \textbf{Latter} & \textbf{Neutral}  \\         \hline
Base vs. w/ clean    &   3.33   &   27.33   &   68.67   \\
Base vs. w/ noisy     &  5.33   &   28.00   &  66.00   \\ 
w/ clean vs. w/ noisy     &  10.67   &  12.67    &  76.67   \\ 
        \hline
Base vs. w/ BERT  &   7.33   &   72.67   &  19.33   \\
Base vs. w/ BERT-CWS+POS   &  8.67   &  78.67   &  12.67  \\
Base vs. w/ BERT-Prosody   &  6.67   &   81.33   &  12.00  \\
Base vs. w/ BERT-Multi     &  4.67    &  85.33    &  9.33   \\
w/ BERT vs. w/ BERT-CWS+POS    &  14.00   &  27.33    &   58.67  \\
w/ BERT vs. w/ BERT-Prosody     &  17.33   &   31.33   &  51.33   \\
w/ BERT vs. w/ BERT-Multi   &  7.33   &  34.67    &   58.00  \\ 

\hline
\multicolumn{4}{r}{Base: FastSpeech 2.  } 
\end{tabular}}
\end{table}

\subsubsection{A case study}
\begin{figure}[htb]
  \centering
    \includegraphics[width=90mm, height=50mm]{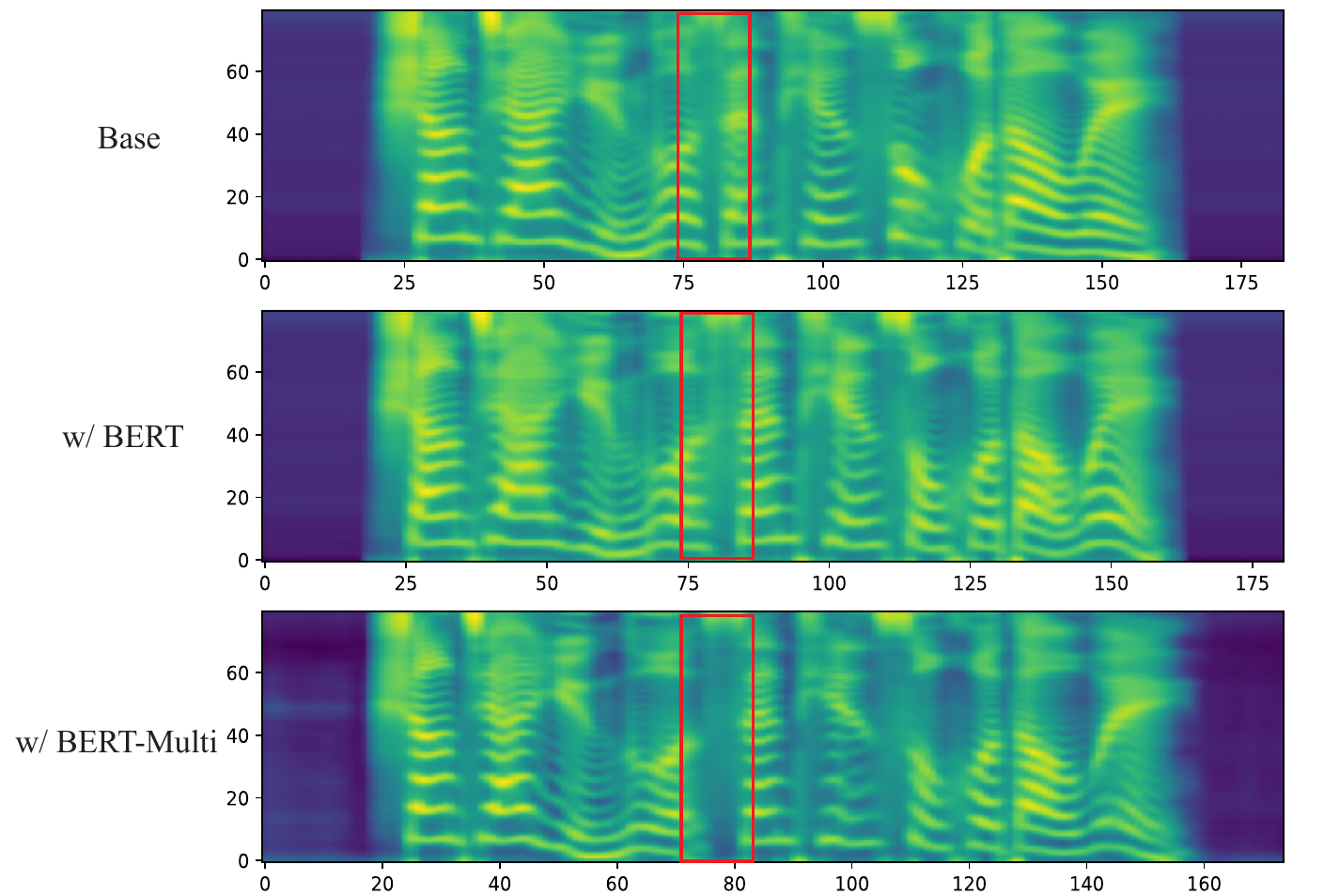}
\caption{Generated mel-spectrograms of the utterance \textbf{shang1 jia1 yun3 nuo4 \#2 san1 tian1 song4 huo4 dao4 wei4} (``The merchant promises that he/she will deliver the goods in three days"). The w/ BERT-Multi architecture synthesizes the most natural break in the prosodic phrase boundary position (\#2).}

\label{fig:demo-mel}
\end{figure}

To investigate if the proposed method can generate better prosody, we select a structurally complex sentence for case study, as shown in Figure \ref{fig:demo-mel}. Based on our linguistic knowledge, there is supposed to be a relatively noticeable pause between the verb of the main clause and the following subordinate clause in the object clause structure. The illustration of generated spectrograms indicates that the w/ BERT-Multi architecture generates speech with more appropriate breaks than the w/ BERT model and FastSpeech 2.

\section{Conclusions}
\label{sec:fin}
In this paper, we propose to improve the prosody for unseen texts by combining a fine-tuned BERT-based front-end with a pre-trained FastSpeech2-based acoustic model. Subjective listening results demonstrate the effectiveness of our proposed methods. Future work will examine the feasibility of training the front-end and the acoustic model in a unified manner, so as to simplify the overall architecture.




\vfill\pagebreak

\bibliographystyle{IEEEbib}
\bibliography{strings,refs}

\end{document}